\documentclass[11pt,a4paper]{article}
\usepackage[hyperref]{emnlp-ijcnlp-2019}
\usepackage{times}
\usepackage{latexsym}
\usepackage{graphicx}
\usepackage{todonotes}
\usepackage{booktabs}
\usepackage{dirtytalk}
\usepackage{microtype}

\usepackage{amsmath}
\usepackage{amssymb}
\usepackage{subcaption}
\usepackage{xspace}
\usepackage{enumitem}

\usepackage{url}

\newcommand{\fastalign}{FastAlign\xspace}
\newcommand{\ench}{En-Zh\xspace}

\title{A Discriminative Neural Model for Cross-Lingual Word Alignment}

\author{Elias Stengel-Eskin\hspace{2em}Tzu-Ray Su\hspace{2em}Matt Post\hspace{2em}Benjamin Van Durme\\
Johns Hopkins University\\
\texttt{\{elias, tsu6\}@jhu.edu}\\
\texttt{\{post, vandurme\}@cs.jhu.edu}
} 

\date{}

\begin{document}
\aclfinalcopy

\maketitle
\begin{abstract}
We introduce a novel discriminative word alignment model, which we integrate into a Transformer-based machine translation model.  In experiments based on a small number of labeled examples ($\sim$1.7K\textendash5K sentences) we evaluate its performance intrinsically on both English-Chinese and English-Arabic alignment, where we achieve major improvements over unsupervised baselines (11\textendash27 F1). We evaluate the model extrinsically on data projection for Chinese NER, showing that our alignments lead to higher performance when used to project NER tags from English to Chinese.
Finally, we perform an ablation analysis and an annotation experiment that jointly support the utility and feasibility of future manual alignment elicitation.

\end{abstract}

\section{Introduction}

Neural-network-based models for Machine Translation (MT) have set new standards for performance, especially
when large amounts of parallel text (bitext) are available. 
However, explicit word-to-word alignments, which were foundational to pre-neural statistical MT (SMT) \cite{brown.p.1993}, have largely been lost in neural MT (NMT) models.
This is unfortunate: while alignments are not necessary for NMT systems, they have a wealth of applications in downstream tasks, such as transferring input formatting, incorporating lexica, and human-in-the-loop translation. 
Crucially, they are central to cross-lingual dataset creation via projection \citep{yarowsky.d.2001}, where token-level annotations in a high-resource language are projected across alignments to a low-resource language; using projection, datasets have been created for a variety of natural language processing tasks, including named-entity recognition (NER), part-of-speech tagging, 
parsing, information extraction (IE), and semantic role labeling \citep{yarowsky.d.2001, hwa.r.2005, riloff.e.2002, pado.s.2009, xi.c.2005}. 
This paradigm allows researchers to address a lack of annotated multilingual resources and changing ontologies with minimal annotation effort. 

\begin{figure}[t]
  \begin{center}
    \includegraphics[width=.48\textwidth]{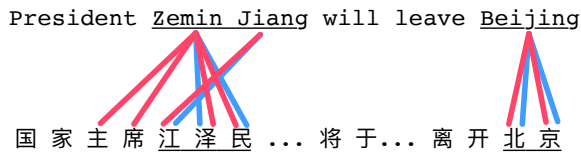}
  \end{center}
  \vspace{-1em}
  \caption{ One application of alignments is to project annotations (e.g., Named Entities) across bitexts.  Here we see a lack of correspondence between \fastalign (red) alignments and gold alignments (blue).\vspace{-2em}}
  \label{fig:example}
\end{figure}

The introduction of attention \citep{bahdanau.d.2014} has allowed NMT to advance past decoding from a single sentence-level vector representation and has been crucial in supporting fluent, meaningful translations, especially of longer sentences.
However, despite their intuitive analogy to word alignments, attention-based alignments typically do not correspond very closely to gold-standard human annotations and suffer from systematic errors \citep{koehn.p.2017, ghader.h.2017}, falling short of the promise to jointly learn to align and translate. 
This problem has only been exacerbated by the introduction of deeper models containing multiple attention layers and heads (e.g. \citet{vaswani.a.2017}).
There have been several attempts to mitigate this discrepancy, mostly with a view to improving the quality of output translations \citep{alkhouli.t.2016, alkhouli.t.2018, chen.w.2016, liu.l.2016} -- though some work has focused specifically on alignment \citep{legrand.j.2016, zenkel.t.2019}. 
We confirm that the poor performance of encoder-decoder attention for Recurrent Neural Networks (RNNs) found by \citet{koehn.p.2017} and \citet{ghader.h.2017} can also be observed in the self-attentional Transformer model \citep{vaswani.a.2017}, further motivating an approach focusing explicitly on alignment quality. 

We introduce a novel alignment module that learns to produce high-quality alignments after training on 1.7K to 4.9K human-annotated alignments for Arabic and Chinese, respectively.
While our module is integrated into the state-of-the-art Transformer model \citep{vaswani.a.2017}, the implementation is architecture-neutral, and can therefore be applied to RNN-based models \citep{schwenk.h.2012, kalchbrenner.j.2013, sutskever.i.2014} or fully convolutional models \citep{gehring.j.2017}. 
Our experiments on English-Chinese and English-Arabic bitext show that our model yields major improvements of 11 and 27 alignment F1 points (respectively) over baseline models; 
ablation experiments indicate with only half the data minor improvements can be observed for Chinese, and that the amount of annotated alignment data has a far greater impact on alignment score than the amount of unlabelled bitext used for pre-training. 

Furthermore, our NER projection trials demonstrate a major downstream improvement when using our alignments over \fastalign. 
Taken together, these results motivate the further annotation of small amounts of quality alignment data in a variety of languages. We demonstrate that such annotation can be performed rapidly by untrained L2 (second language) speakers.

\section{Related work}
\paragraph{Statistical Alignment}
Generative alignment \citep{brown.p.1993} models the posterior over a target sequence $\mathbf{t}$ given a source sequence $\mathbf{s}$ as:
\begin{align}
    \nonumber
    p(\mathbf{t} | \mathbf{s}) = 
    &\sum_a \prod_i 
    \underbrace{p(t_i | a_i, \mathbf{t}_{<i}, \mathbf{s})}_\text{lexical model} \underbrace{p(a_i | \mathbf{a}_{<i}, \mathbf{t}_{<i}, \mathbf{s})}_\text{alignment model}
\end{align}
\vspace{-4mm}

Such models are asymmetrical: they are learned in both directions and then heuristically combined. 

Discriminative alignment models, on the other hand, directly model $p(\mathbf{a} | \mathbf{s}, \mathbf{t})$, usually by extracting features from the source and target sequences and training a supervised classifier using labelled data.
A comprehensive account of methods and features for discriminative alignment can be found in \citet{tomeh.n.2012}. 

\paragraph{Neural Machine Translation}
The attentional encoder-decoder proposed by \citet{bahdanau.d.2014} greatly improved the initial sequence-to-sequence architecture; rather than conditioning each target word on the final hidden unit of the encoder, each target word prediction is conditioned on a weighted average of all the encoder hidden units.
However, the weights in the average do not constitute a soft alignment, as confirmed by \citet{koehn.p.2017} and \citet{ghader.h.2017}.
Our findings align with those of \citet{ghader.h.2017}, who provide a detailed analysis of where attention and alignment diverge, offering possible explanations for the modest results obtained when attempting to bias attention weights towards alignments.
For example, \citet{chen.w.2016} introduce a loss over the attention weight that penalizes the model when the attention deviates from alignments; this yields only minor BLEU score improvements.\footnote{This may also be a result of noisy training data, as they use unsupervised alignments as training data for their models.}

\paragraph{Neural Alignment Models} 
\citet{legrand.j.2016} develop a neural alignment model that uses a convolutional encoder for the source and target sequences and a negative-sampling-based objective. 
\citet{tamura.a.2014} introduce a supervised RNN-based aligner which conditions each alignment decision on the source and target sequences as well as previous alignment decisions.
\citet{alkhouli.t.2018} extend the  Transformer architecture for alignment by adding an additional alignment head to the multi-head encoder-decoder attention, biasing the attention weights to correspond with alignments.
Similarly, \citet{zenkel.t.2019} introduce a single-layer attention module (akin to the standard Transformer decoder) which predicts the target word conditioned on the encoder and decoder states, following the intuition that a source word should be aligned to a target word if it is highly predictive of the target word. 
Unlike our model, all of these models are asymmetrical and are trained without explicit alignment data. 

Following the generative formulation of the alignment problem, \citet{alkhouli.t.2016} present neural lexical translation and alignment models, which they train using silver-standard alignments obtained from GIZA++. 
A similar training strategy is used in \citet{alkhouli.t.2017}, who bias an RNN attention module with silver-standard alignments for use as a lexical model. 
In the same vein, \citet{peter.j.2017} improve the attention module by allowing it to peek at the next target word. 

Note that unlike our framework, none of the models mentioned make use of existing gold-standard labeled data. Following \citet{legrand.j.2016} and \citet{zenkel.t.2019} we focus exclusively on alignment quality. A similar approach is taken by \citet{ouyang.j.2019}, who introduce a pointer network-based model for monolingual phrase alignment which allows for the alignment of variable-length phrases.


\section{Model}

NMT systems typically use an encoder-decoder architecture, where one neural network (the encoder) learns to provide another network (the decoder) with a continuous representation of the input sequence of words in the source language, $s_1, \ldots, s_N$,  from which the decoder can accurately predict the target sequence $t_1, \ldots, t_M$. 
The intermediate representations of the encoder and decoder are known as hidden states.
After training, the hidden states of the encoder ($S_{1}, \ldots, S_{N}$) and decoder ($T_{1}, \ldots, T_{M}$) contain information about their co-indexed source and target words as well as their respective contexts; this can be thought of as a rough approximation of the semantic content of the source and target words. 
In an MT model (such as the Transformer-based model used here) the states of the encoder and decoder networks also bear information on cross-lingual correspondence. 

From these representations, we can derive a natural formulation of the alignment problem: words of the source language should be aligned to the words in the target language that convey roughly the same meaning in the translated sentence i.e. their contextual vector-based meaning representations are similar.

We use the dot product as an unnormalized measure of distance between each pair of vectors $S_i$ and $T_j$. In order to apply a distance metric, the two vectors are projected into a shared space via a shared three-layer feed-forward neural network with \emph{tanh} activations and a linear output layer, producing source and target matrices $S'_{1},\ldots, S'_{N}$ and $T'_{1}, \ldots, T'_{M}$.\footnote{We use 512 hidden units in each layer.} This is given by:

\vspace{-7mm}
\begin{align}
\nonumber S'_i &= W_{3}(tanh(W_{2}(tanh(W_{1} S_i) )) ) \\
\nonumber T'_j &= W_{3}(tanh(W_{2}(tanh(W_{1} T_j) )) )
\end{align}
\vspace{-6mm}

Taking the matrix product of the projected source and target sequences along their shared dimension yields a matrix A of $N \times M$ semantic ``links'', where each element $A_{ij}$ is an un-normalized distance between vectors $S'_{i}$ and $T'_{j}$, i.e. 
$ A = [S'_1, \ldots, S'_N] \cdot [T'_1, \ldots, T'_M]^T $

These distances will be passed through a sigmoid function, normalizing them to probabilities $p(a_{ij} | \mathbf{s}, \mathbf{t})$. However, alignments are context-sensitive: the decision to align source word $i$ to target word $j$ may affect the probability of aligning the neighbors of $i$ and $j$. Therefore, before applying the sigmoid function to the matrix $A$ we convolve it with a shared weight matrix, yielding a matrix $A'$, which encodes the relationships between alignments; this is given by 
$A' = W_{conv} \ast A $, where $\ast$ is the convolution operator. This can be thought of as a way of conditioning each alignment decision on its neighbors, modeling $p(a_{i,j} | \mathbf{s}, \mathbf{t}, a_{i'\neq i,j'\neq j})$. This step is crucial\textemdash without a $3\times 3$ convolution, F1 results hover below 50. 

Finally, we apply the sigmoid function to $A'$ and treat each alignment decision as a separate binary classification problem, using binary cross entropy loss:

\vspace{-6mm}
\begin{align}
    \nonumber
    \sum\limits_{i < N, j < M} \Big(& \hat{a}_{ij} \log \big(p(a_{ij} | \mathbf{s}, \mathbf{t}) \big) \:\: +  \\[-1em]
    \nonumber
    &\big(1-\hat{a}_{ij}\big)\log\big(1 - p(a_{ij} | \mathbf{s}, \mathbf{t})\big) \Big)
\end{align} 
\vspace{-6mm}

\noindent where $\hat{a}_{ij} =1 $ if source word $i$ is aligned to target word $j$ in the gold-standard data, and $0$ otherwise. Note that the convolution step ensures these are not independent alignment decisions. 

When training our model, we begin by pre-training an MT model on unlabelled bitext; the weights of this model are used to initialize the encoder and decoder in the alignment model. 
To obtain an alignment at test time, the source and target word sequences are encoded and presented to the aligner.

\section{Alignment Experiments}
\label{sec:align}
\subsection{Baselines}

We compare against two baselines: alignments obtained from FastAlign~\citep{dyer.c.2013}, and attention-based alignments. FastAlign is a fast log-linear reparameterization of IBM Model 2 \citep{brown.p.1993}.

Unlike in the RNN case, where a single attention layer intervenes between the encoder and decoder, Transformers make use of multi-head and multi-layer attention. Our attentional alignments are obtained by averaging across all heads of the final multi-head encoder-decoder attention layer.

Both of these methods allow for fast online decoding, making them comparable to our model, and both have been used for word alignment; moreover, \fastalign is a standard choice for projection  \citep{arthur.p.2016, fu.r.2014, eger.s.2018, agic.z.2016, sharoff.s.2018}.

Our model extends the Sockeye implementation of the Transformer model \citep{hieber.f.2017}.\footnote{We pre-train a 6-layer model with the Adam optimizer \citep{kingma.d.2014} using a learning rate of $0.0002$, 8-headed multi-head attention, 2048-dimensional feed-forward layers, and 512-dimensional encoder/decoder output layers.
We include weight-tying \citep{press.o.2017} between the source and target embedding layers.
All other MT hyperparameters were set to the Sockeye defaults.}

Following \citet{dyer.c.2013}, \fastalign hyperparameters were tuned by training on 10K training examples, evaluating on a 100-sentence subset of the validation split. The \emph{grow-diag-final-and} heuristic was used for symmetrization. Thresholds for both attention alignment and our model were tuned on the same 100-sentence validation subset.

\subsection{Data}
Chinese character-tokenized bitext and alignments were sourced from the GALE Chinese-English Parallel Aligned Treebank \citep{li.x.2015}; Arabic data was obtained from the GALE Arabic-English Parallel Aligned Treebank \citep{li.x.2013}.
Both GALE datasets are based on broadcast news corpora; gold-standard alignments were produced by multiple trained human annotators, with several checks for quality including redundant annotation, review by senior annotators, and automatic heuristics. 
The GALE corpora satisfy the following criteria:
\begin{enumerate}[topsep=0pt,itemsep=-1ex,partopsep=1ex,parsep=1ex]
    \item \emph{Quality}: although some problems were detected in the GALE alignments (e.g. mismatched bitext, formatting errors, etc.), the multiple levels of quality control suggest a high level of overall quality.  
    \item \emph{Size}: while other gold-standard alignment corpora exist (Europarl, etc.) these are typically on the order of $\sim$500 sentences; conversely, GALE is large enough to be used as a training set in a neural setting. 
    \item \emph{Annotations}: GALE is annotated with a variety of labels, and a subset of the Chinese portion of GALE was annotated for NER as part of the OntoNotes corpus \citep{weischedel.r.2013}. 
\end{enumerate}
Sentences were  split into train, test, and validation portions and cleaned by removing extraneous annotations (e.g. timestamp tags, etc.). Corpus statistics are given in Table \ref{tab:gale}.

\begin{table}[h]
    \centering
    \begin{tabular}{|l|ccc|}
        \hline
         Language & train & dev & test \\
         \hline
         \hline
         AR & 1687 & 299 & 315 \\
         ZH & 4871 & 596 & 636 \\
        \hline
    \end{tabular}
    \caption{Num. annotated alignment sentence pairs.\vspace{-1em}}
    \label{tab:gale}
\end{table}

The discriminative models (referred to as DiscAlign) were initialized with weights from an MT model pre-trained on 4 million sentences, which were tokenized using the Moses tokenizer \citep{koehn.p.2007}. 
Chinese data was drawn from the WMT 2017 \ench bitext, while the Arabic data was sourced from local resources; both Chinese and Arabic models were trained with a 60K-word vocabulary. For \fastalign, the GALE train split was concatenated to the pretraining bitext, while for both the discriminative and attentional models the GALE data was used to finetune a model pre-trained on bitext alone. 

Contemporary MT systems use subword units to tackle the problem of out-of-vocabulary low-frequency words. 
This is often implemented by byte-pair encoding (BPE) \citep{sennrich.r.2016}. 
We applied a BPE model with 30K merge operations to all of our data for the BPE experimental condition.
Training alignments were expanded by aligning all subwords of a given source word to all the subwords of the target word to which the source word aligns. At test time, following \citet{zenkel.t.2019}, to reduce BPE alignments to regular alignments, we considered a source and target word to be aligned if \emph{any} of their subwords were aligned.\footnote{Note that this step is necessary in order to make the produced and reference alignments comparable; to evaluate alignments, the sequences must be identical, or the evaluation is ill-posed.}
This heuristic was used to evaluate all BPE models. 

With the task of projection for NER data in mind  \citep{yarowsky.d.2001}, we evaluate all models on Chinese NER spans. 
These spans were obtained from the OntoNotes corpus, which subsumes the GALE Chinese corpus. 
Due to formatting differences, only a subset of the GALE sentences (287 validation, 189 test) were recoverable from OntoNotes.

We evaluate our models using macro-F1 score, as \citet{fraser.a.2007} showed that alignment error rate does not match human judgments.
F1 scores were obtained using the macro-F1 scorer provided by FastAlign.

\begin{figure}[t]
    \raggedleft
    \hspace{-10mm} 
    \vspace{-2mm}\includegraphics[width=.5\textwidth]{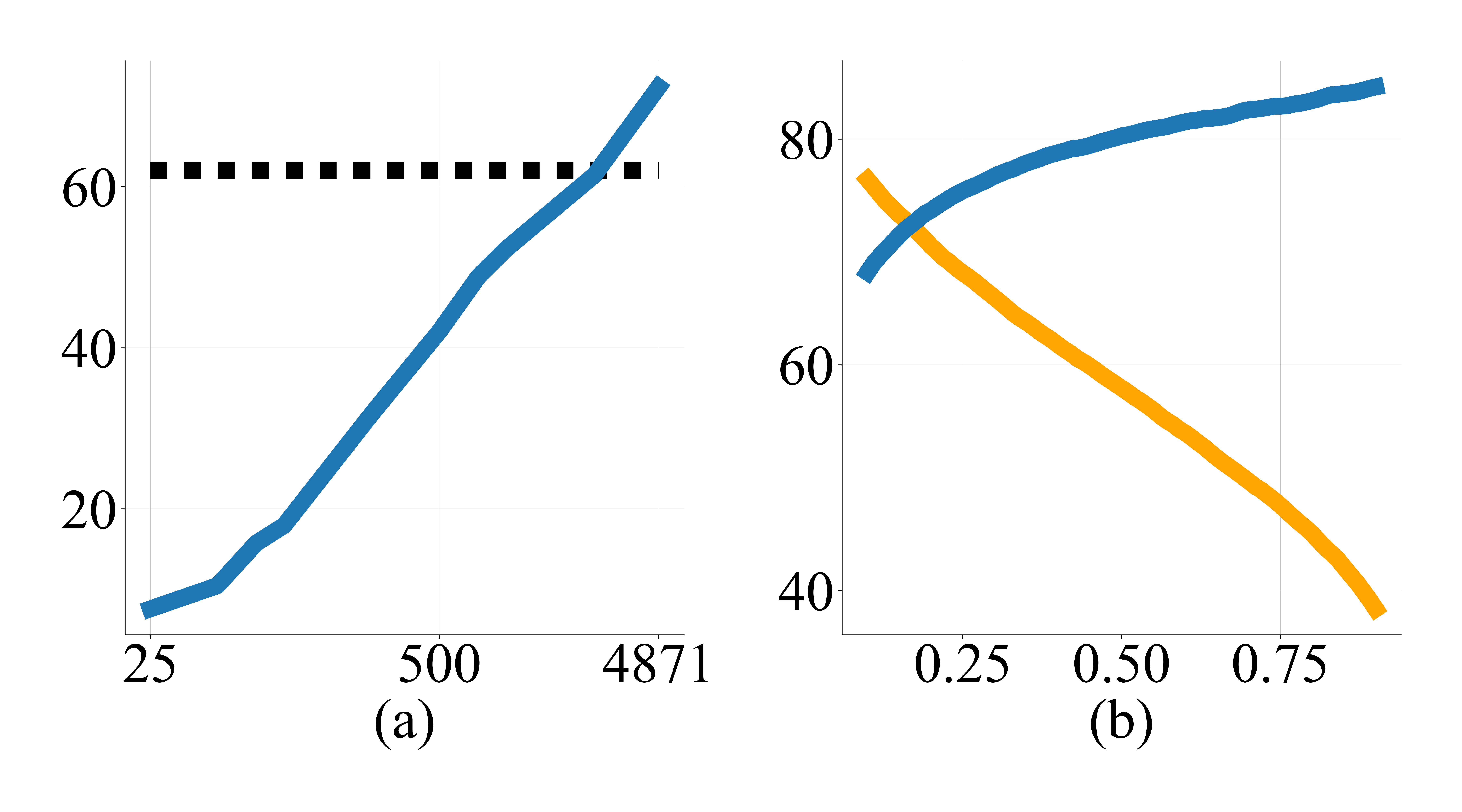}\vspace{-2mm}
\caption{(a) Dev. F1 as function of training corpus size (log scale), contrast against \fastalign F1 (dashed). (b) Dev. precision (blue) and recall (orange) at different thresholds. \vspace{-1.2em}}
\label{fig:combined}
\end{figure}

\begin{table}[t!]
\begin{center}
\begin{tabular}{l | c c | c }
\hline
Method & P & R & F1 \\
\hline
\hline
Avg. attention. & 36.30 & 46.17 & 40.65\\
Avg. attention (BPE) & 37.89 & 49.82 & 43.05 \\
Avg. attention (NER) & 16.57 & 35.85 & 22.66 \\
\hline

FastAlign & 80.46 & 50.46 & 62.02\\
FastAlign~(BPE) & 70.41 & 55.43 & 62.03  \\
FastAlign (NER) & 83.70 & 49.54 & 62.24 \\
\hline
\hline
DiscAlign &  72.92 &  73.91 & \textbf{73.41}  \\
DiscAlign (BPE) & 69.36 & 67.11 & \textbf{68.22}  \\
DiscAlign (NER) & 74.52 & 77.05 & \textbf{75.78} \\
DiscAlign (NER) +prec. & 84.69 & 58.41 & 69.14\\

\hline
\end{tabular}
\end{center}
\caption{\label{tab:zh_res} Precision, Recall, and F1 on Chinese GALE test data. BPE indicates ``with BPE", and NER denotes restriction to NER spans.}
\end{table}

\begin{table}[t!]
\begin{center}
\begin{tabular}{l | c c | c }
\hline
Method & P & R & F1 \\
\hline
\hline
Avg. attention. & 8.46 & 32.50 & 13.42 \\
Avg. attention (BPE) & 10.11 & 17.27 & 12.75 \\
\hline
FastAlign & 62.26 & 51.06 & 56.11 \\
FastAlign~(BPE) & 62.74 & 51.25 & 56.42 \\
\hline
\hline
DiscAlign &  91.30 & 75.66 & \textbf{82.74} \\
DiscAlign (BPE) & 87.05 & 76.98 & \textbf{81.71}  \\
\hline

\end{tabular}
\end{center}
\caption{\label{tab:ar_res} Precision, Recall, and F1 on Arabic GALE test data.\vspace{-1em}}
\end{table}

\subsection {Results} Our model outperforms both baselines for all languages and experimental conditions. Table \ref{tab:zh_res} shows that our model performs especially well in the NER setting, where we observe the largest improvement over \fastalign ($\Delta=14$). By increasing the threshold $\alpha$ above which  $p(a_{i,j} | \mathbf{s}, \mathbf{t})$ is considered an alignment we can obtain high-precision alignments, exceeding \fastalign's precision and recall. The best threshold values on the development set were $\alpha = 0.13$ and $\alpha = 0.14$ for average attention on data with and without BPE (respectively) and $\alpha = 0.15$ for the discriminative settings ($\alpha = 0.5$ was used for the high-precision ``+prec'' setting). 

Table \ref{tab:ar_res} further underscores these findings; we observe an even more dramatic increase in performance on Arabic, despite the fact that the model is trained on only 1.7K labelled examples. Furthermore, we see that average attention performs abysmally in Arabic. The best thresholds for Arabic differed substantially from those for Chinese: for average attention (with and without BPE) the values were $\alpha = 0.05$ and $\alpha = 0.1$, while for the discriminative aligner they were $\alpha = 0.94$ and $\alpha = 0.99$ (compared to $0.15$ for Chinese). This is reflected in the high precision of the discriminative aligner. 

Figure \ref{fig:combined}(a) shows the relationship between the number of training examples and validation F1 performance for Chinese, with the performance of FastAlign overlaid as a benchmark. We observe that with a little over half the training examples (2500), we perform comparably to FastAlign. 

\begin{figure}[t]
    \raggedleft
    \includegraphics[width=.5\textwidth]{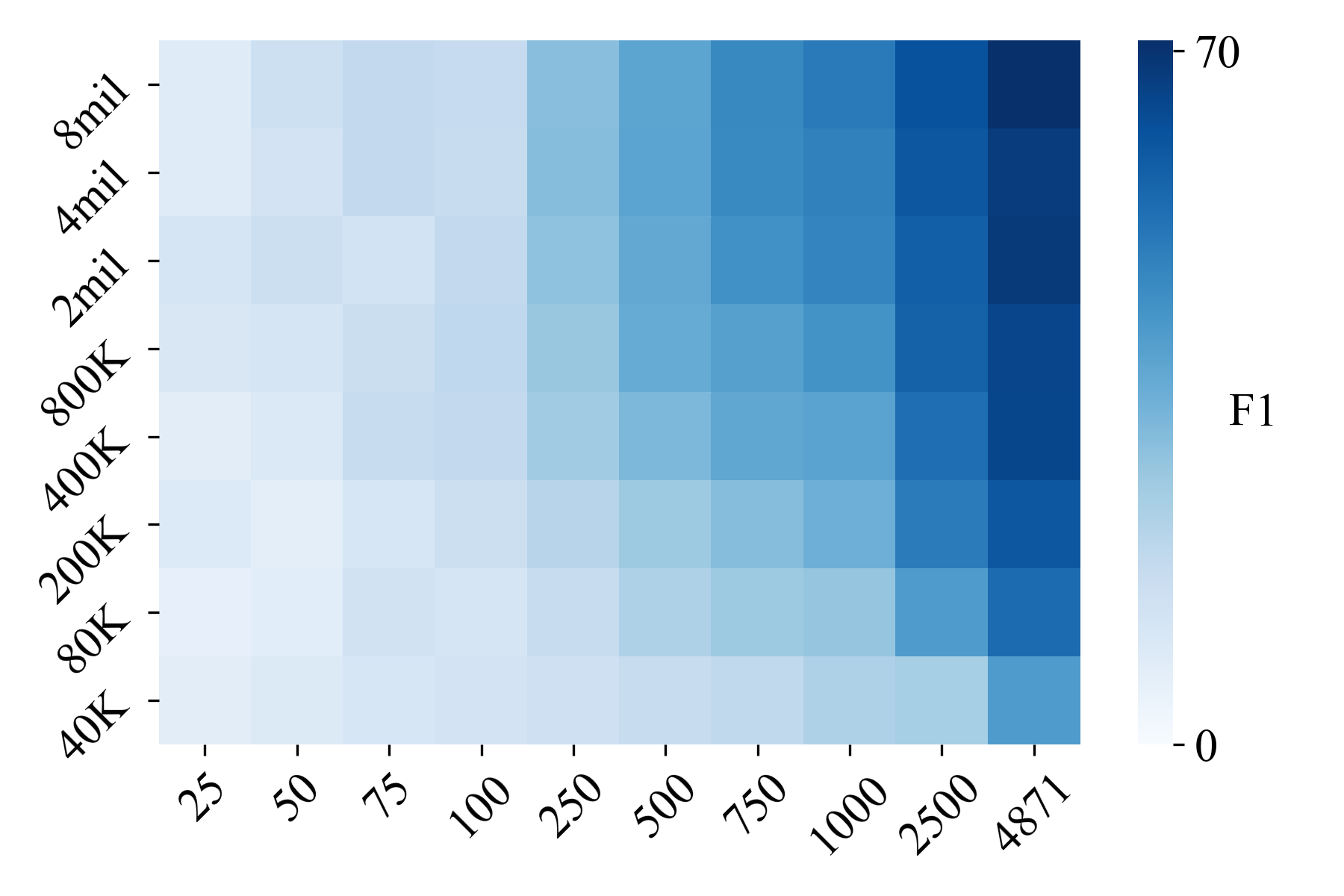}
\caption{Chinese F1 performance with respect to the amount of unlabelled pretraining bitext (y axis) and the number of sentences annotated for alignment (x axis); increasing the number of annotated sentences has a far greater impact on the F1 score than increasing the amount of bitext.\vspace{-1.5em}}
\label{fig:heatmap}

\end{figure}

In Figure \ref{fig:heatmap}, we demonstrate that labelled alignment data has a greater impact than the amount of pre-training bitext. For example, halving the number of pre-training examples from 8 million to 4 million only reduces F1 by 3.69, but halving the number of labelled finetuning examples while keeping the number of pre-training sentences constant at 8 million lowers the performance by 9.44 F1.

\paragraph{Analysis} Taken together, Table \ref{tab:zh_res} and Figure \ref{fig:combined} support the use of discriminative alignment models in settings where test data is presented in an online fashion as well as in cases where high-quality alignments matter. Most importantly, the results in the NER condition highlight the applicability of our model to dataset projection for NER.

Figure~\ref{fig:combined}(b) also shows that by varying the minimum threshold $\alpha$ on $p(a_{ij} | \textbf{s}, \textbf{t})$, we can manipulate the precision-recall tradeoff, giving our model an advantage in scenarios where high precision alignments are required (e.g., dataset projection) as well as in ones where high recall is preferred (e.g., correction by human annotators). 

Our results underscore the poor performance of attentional alignments. 
One factor driving the low F1 scores observed may be the Transformer's design: each attention head in the Transformer's multi-head attention layer learns to attend to different aspects of the encoder states; thus, averaging across all heads is likely to yield even noisier results than the RNN model, where there is effectively only one attention head. 
Another reason attention performs poorly may be that attention heads are learning task- and dataset- specific patterns that do not match alignments but nevertheless provide useful cues for translation.
This would support the observations of \citet{ghader.h.2017}, who find that for certain parts of speech (e.g. verbs) alignment is very poorly correlated with translation loss. 

 These findings illustrate the benefit of supervised training for alignment, suggesting further annotation of gold-standard alignments is warranted, with an emphasis on alignment quality.\footnote{In addition to the preprocessing steps required to clean the data, a bilingual speaker pointed out several systematic deviations from the alignment protocol by GALE annotators.} In particular, Figure~\ref{fig:heatmap} indicates that labelled alignment data is far more valuable than bitext, meaning that efforts should be focused on annotating existing bitext for alignment; \fastalign's performance (when trained on the entire 4 million-sentence training bitext and the alignment bitext) can be surpassed after pre-training the translation model on only 400K sentences of unlabelled bitext and the full alignment set (4871 sentences). 

\section{NER Experiments}
\label{sec:ner}
In Table \ref{tab:zh_res} we evaluated the quality of our model's alignments intrinsically, showing that they dramatically outperform the baseline models, especially for NER spans. 
To show that these improvements translate to downstream performance, we conduct projection experiments for NER using the Chinese and English data from the OntoNotes 5.0 corpus.

\paragraph{Projection} 
Dataset projection, as defined by \citet{yarowsky.d.2001}, is a method of obtaining data in a low resource language by leveraging existing bitext and high-resource annotations. 
As shown (for a single sentence) in Figure~\ref{fig:example}, a set of sentences in a high-resource language for which there exist task-specific token-level annotations (e.g. an English NER dataset) is translated into the target language. 
Using alignments, the annotations are ``projected'' from the source text to the target; each target token receives the label of the source token it is aligned to (or a default label if the token is unaligned). 
This produces an annotated resource in the target language which can be used to train or pre-train a model.

\paragraph{Data} We project the English OntoNotes NER training data (53K sentences) by first translating it to Chinese using our BPE translation model, and then aligning the source and target sequences using \fastalign and our best discriminative model. We create two datasets, one where the tags have been projected using \fastalign alignments and another using the higher-quality DiscAlign alignments. OntoNotes contains gold-labelled NER data for 18 tag types in both Chinese and English. The official splits\footnote{\url{http://conll.cemantix.org/2012/data.html}} were used. Because there are fewer training examples for Chinese than English, two NER models were trained, one with the same number of projected sentences as there were gold Chinese sentences (36K), and the second using the full English dataset (53K sentences) projected to Chinese. 
The data was pre-processed in following the same procedure as \S \ref{sec:align}

\paragraph{NER Model} Given the recent successes of contextual encoders for sequence modeling, especially for NER, we make use of a pre-trained BERT model \citep{devlin.j.2018}, where the NER label is determined by the softmax output of a linear layer situated on top of the encoder.\footnote{An NER model was adapted from \url{http://github.com/kamalkraj/BERT-NER} to use a pre-trained Chinese model and allow for incremental checkpointing.} Each model was trained for 4 to 8 epochs. 

We explore the performance of the BERT NER model when trained on both gold-standard and projected datasets. In addition, we analyze its performance when trained on decreasing amounts of gold data, and contrast those results with the performance of a model pre-trained on projected data and fine-tuned with gold data.

\subsection{Results \& Analysis}

\begin{table}[t!]
\begin{center}
\begin{tabular}{l | c | c c | c }
\hline
Method & \# train & P & R & F1 \\
\hline
\hline
Zh Gold & 36K & 75.46 & 80.55 & 77.81 \\
\hline
\fastalign & 36K & 38.99 & 36.61 & 37.55 \\
\fastalign & 53K & 39.46 & 36.65 & 37.77 \\
DiscAlign & 36K & 51.94 & 52.37 & 51.76 \\
DiscAlign & 53K & 51.92 & 51.93 & 51.57 \\
\hline
\end{tabular}
\end{center}
\caption{\label{tab:ner_res} F1 results on OntoNotes test for systems trained on data projected via \fastalign and DiscAlign. \vspace{-1.5em}}

\end{table}

\begin{table}[t!]
\begin{center}
\begin{tabular}{l | c c | c }
\hline
\# train & P & R & F1 \\
\hline
\hline
500 & 41.43 & 57.97 & 48.07 \\ 
1000 & 58.52 & 68.05 & 62.72 \\
2500 & 68.30 & 74.93 & 71.35 \\
4871 & 72.00 & 76.90 & 74.30 \\
10K & 73.13 & 78.79 & 75.77 \\
\hline
\end{tabular}
\end{center}
\caption{\label{tab:ner_abl} OntoNotes test set performance when trained on subsamples of the Chinese gold NER data. \vspace{-0.5em}}

\end{table}

\begin{table}[t!]
\begin{center}
\begin{tabular}{l | c c | c | c }
\hline
 \# gold & P & R & F1 & $\Delta$F1 \\
\hline
\hline
500 & 63.16 & 68.90 & 65.74 & 17.67 \\ 
1000 & 66.39 & 72.23 & 69.13 &  6.41 \\
2500 & 69.57 & 75.52 & 72.33 & 0.98 \\
4871 & 71.11 & 76.38 & 73.60 & -0.7 \\
10K & 72.52 & 78.32 & 75.24 & -0.53\\
full & 75.04 & 80.18 & 77.45 & -0.36\\
\hline
\end{tabular}
\end{center}
\caption{\label{tab:ner_pre} OntoNotes test performance when pre-trained on projected data and finetuned on varying amounts of Chinese gold NER data. $\Delta $ F1 compares these values to corresponding rows in Table \ref{tab:ner_abl}. \vspace{-1.0em}}

\end{table}

\begin{figure*}[t]
    \raggedleft
    \includegraphics[width=\textwidth]{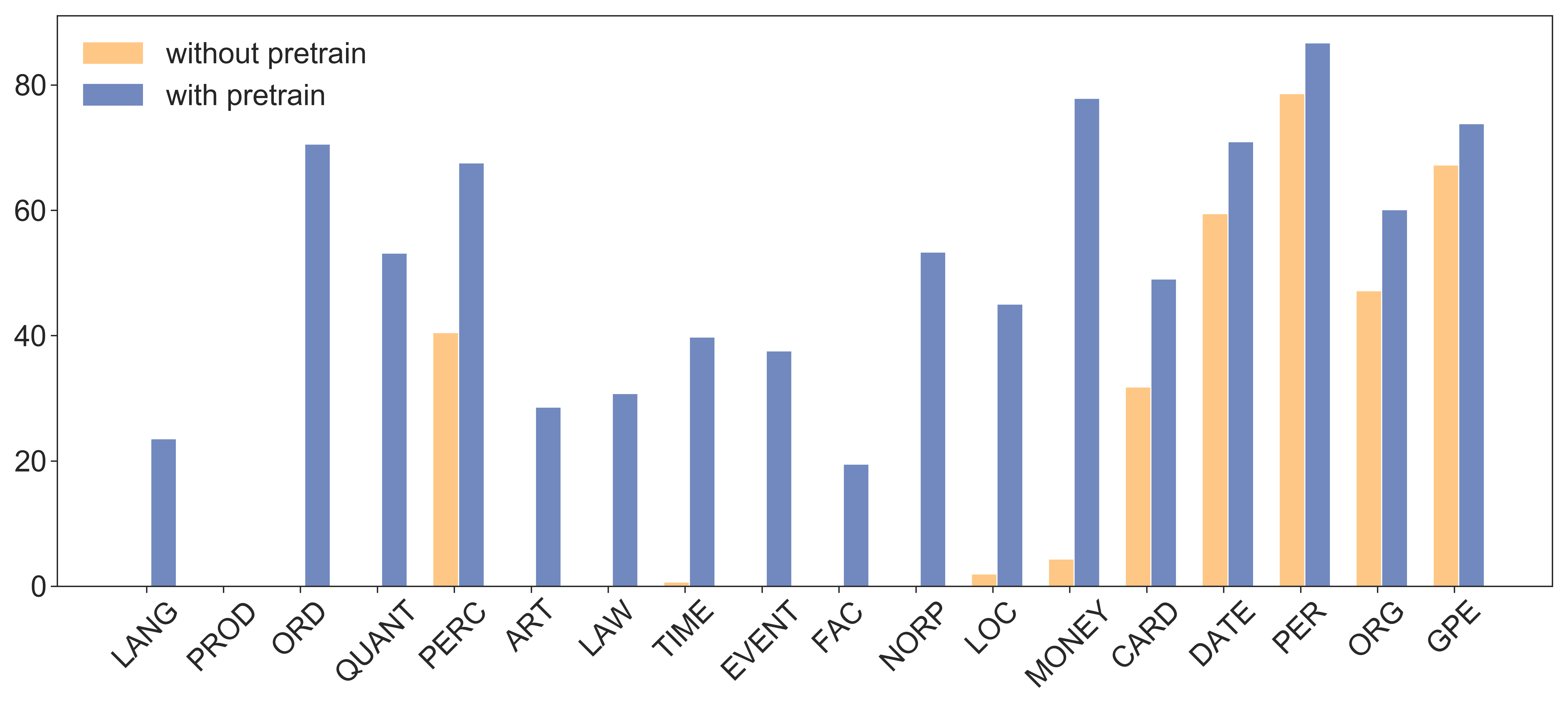}
\caption{Comparing F1 across different tag types when a model is trained on 500 gold NER sentences to when it is pre-trained on projected data and then finetuned on 500 gold sentences, ranked by number of train examples. \vspace{-1em}}
\label{fig:classes}

\end{figure*}

Table \ref{tab:ner_res} shows that while NER systems trained on projected data do categorically worse than an NER system trained on gold-standard data, the higher-quality alignments obtained from DiscAlign lead to a major improvement in F1 when compared to \fastalign. 

For many tasks, an English resource (i.e. the English portion of the OntoNotes data) exists along with bitext between English and a given target language, but very few resources are available in the target language. Table \ref{tab:ner_abl} shows that projection outperforms a system trained on gold data when there is very little gold data available (500 sentences); as the amount of data increases, these gains disappear. However, Table \ref{tab:ner_pre} indicates that projected data provides a useful pre-training objective for NER, making low-data NER models perform about as well as if they had been trained on twice as much gold data. 

Figure \ref{fig:classes} shows that for nearly all classes, especially rare ones, pre-training on projected data improves downstream results when there is little annotated data. 
This is partly due to the fact some classes (e.g. LANGUAGE, ORDINAL, etc.) may appear very rarely in the 500-sentence training set ($\leq 21$ times). 
Note that all classes appear at least 5 times in the training data. 
However, these classes are far more common in the much larger projected dataset. 
In addition, the major gains seen in tags with higher support in the gold data (e.g. MONEY, LOCATION) suggest that pre-training the NER model on projected data does more than merely expose the model to unseen classes. 

\section{Human Evaluation}

In \S \ref{sec:align} we established that our  discriminatively-trained neural aligner outperforms unsupervised baselines, especially on NER spans; in \S \ref{sec:ner} we verified that the alignments it produces can be productively applied a downstream IE task (NER) via dataset projection. 

However, unlike these unsupervised baselines, our aligner requires labelled data on the order of thousands of sentences, and thus cannot be applied to language pairs for which no labelled alignment data exists (most languages). 
We contend that using a web-based crowdsourcing interface \footnote{\url{https://github.com/hltcoe/tasa}}, alignment annotation can be performed rapidly by annotators with minimal experience; to justify this claim, we conduct a small timed experiment using untrained human annotators.
Six L2 Chinese speakers (L1 English) were asked to annotate English-Chinese bitext for alignment. 
These sentences were taken from the GALE development portion of the OntoNotes dataset. 
Note that these annotators had no experience in annotating data for alignment, and had approximately equivalent amounts of Chinese instruction. 
Annotators were introduced to the task and the alignment interface, and then given 50 minutes to align the sentences as accurately as possible. 
They were divided into two groups, with each group annotating a different partition of sentences.

\begin{table}[t!]
\begin{center}
\begin{tabular}{l | c | c c | c }
\hline
Model & sents/m & P & R & F1 \\
\hline
\hline

Human  & 4.4 & 90.09 & 62.85 & 73.92 \\
DiscAlign  & - & 74.54 & 72.15 & 73.31 \\
\hline
Hu. (NER) & - & 87.73 & 71.02 & 78.24 \\
DA (NER) & - & 77.37 & 67.69 & 71.94 \\
\hline
\end{tabular}
\end{center}
\caption{\label{tab:aln_time} Sentences per minute and average scores against gold-labelled data for sentences annotated for alignment by human annotators (Hu.), compared to DiscAlign (DA) on the same sentences. \vspace{-1.5em}}

\end{table}

\subsection{Results \& Analysis} Table \ref{tab:aln_time} shows that alignment can be performed rapidly, at 4.4 sentences per minute. 
The annotators achieve high overall precision for alignment, but fall short on recall. 
When only alignments of NER spans are considered, their F1 score improves considerably.
Additionally, human annotators outperform DiscAlign when evaluated on the same sentences. 

These are promising results, as they indicate that entirely untrained L2 annotators can rapidly annotate data. 
Their relatively low recall scores might be due to their lack of training, and given more extensive training and guidelines we would expect them to improve.

\section{Conclusion \& Future Work}
We described a model extension to NMT which supports discriminative alignment, where fewer than 5K Chinese and 1.7K Arabic labelled sentences lead to better F1 scores in all tested scenarios than standardly employed mechanisms. By evaluating the model on both Arabic and Chinese we demonstrated that our model's improvements generalize across typologically divergent languages. We showed that in Chinese our model's performance is particularly strong for NER spans. 

We projected a dataset for NER from English to Chinese using our alignments, verifying that our model's higher alignment quality led to downstream improvements; our aligner's intrinsic improvements result in major gains over \fastalign in a downstream NER evaluation. 
Comparing the performance of our projected dataset with a gold-standard dataset, we showed that a large projected dataset can outperform a small labelled one, and that projection can be leveraged as a method of data augmentation for low-resource settings. 

Finally, we showed that untrained L2 annotators can rapidly annotate bitext for alignment, and yield promising results for both full alignment and NER-specific annotation, confirming that we can collect additional data in a variety of languages.  

Based on our findings that:
\begin{itemize}[topsep=0.0pt,itemsep=-1ex,partopsep=1ex,parsep=1ex]
    \item discriminative alignments outperform common unsupervised baselines in two typologically divergent languages
    \item this performance boost leads to major downstream improvements on NER
    \item only a small amount of labelled data is needed to realize these improvements 
    \item and that these labelled examples can be obtained from L2 speakers with minimal training 
\end{itemize}
we conclude with a call for additional annotation efforts in a wider variety of languages. While multilingual datasets directly annotated for a given task typically lead to the highest-performing systems (see Table \ref{tab:ner_res}) these datasets are task-specific and not robust to ontology changes. In contrast, the framework of projection via discriminatively-trained alignments which we present is task-agnostic (any token-level annotations can be projected) and requires only one source-side dataset to be annotated. 

Future work will explore different sets of languages paired with new, higher-quality elicited alignments of existing bitext. 
We will also explore architecture modifications to allow for task-specific annotation, training, and inference for a given IE task, where only certain spans are of interest (i.e. training our aligner only on NER-span alignments). We expect that asking annotators to only align Named Entities (or some other token type) will greatly increase the annotation speed. These experiments will extend beyond NER to other token-level tasks such as coreference resolution and event detection.

\section*{Acknowledgements}
This work was supported by DARPA LORELEI and AIDA. We thank the anonymous reviewers for their constructive feedback. The views and conclusions expressed herein are those of the authors and should not be interpreted as representing official policies or endorsements of DARPA or the U.S. Government.

\bibliography{alignment}
\bibliographystyle{acl_natbib}
\end{document}